\documentclass[letterpaper, 10pt, conference]{ieeeconf}  

\IEEEoverridecommandlockouts                              

\usepackage{graphics} 
\usepackage{epsfig} 
\usepackage{times} 
\usepackage{amsmath} 
\usepackage{amssymb}  
\usepackage[ruled,noend,vlined,linesnumbered]{algorithm2e}
\usepackage{xcolor}

\usepackage{amsthm}
\usepackage{multirow}
\usepackage{subcaption}
\usepackage[font=footnotesize,labelfont=bf]{caption}
\let\labelindent\relax
\usepackage{enumitem}
\usepackage{titlesec}   

\theoremstyle{remark}
\newtheorem*{remark}{Remark}

\renewcommand{\tilde}{\widetilde}
\newcommand{\E}{\mathbb{E}}
\newcommand{\R}{\mathbb{R}}

\newcommand{\mA}{\mathcal{A}}

\newcommand{\mQ}{\mathcal{Q}}
\newcommand{\mS}{\mathcal{S}}

\newcommand{\1}{\mathbf{1}}
\newcommand{\abs}[1]{\left \vert #1 \right\vert}

\newcommand{\tp}{\mathsf{T}}
\SetKw{KwInit}{Initialize:}
\SetKw{kWReturn}{Return:}
\SetKw{kWBreak}{break}

\titlespacing\section{0pt}{6pt plus 3pt minus 2pt}{2pt plus 3pt minus 2pt}
\titlespacing\subsection{0pt}{4pt plus 3pt minus 2pt}{2pt plus 3pt minus 2pt}
\title{\LARGE \bf 
Stackelberg Strategic Guidance for Heterogeneous Robots Collaboration 
}

\author{Yuhan Zhao,\;\; Baichuan Huang,\;\; Jingjin Yu\; and\; Quanyan Zhu 
\thanks{Y. Zhao and Q. Zhu are with the Department of 
Electrical and Computer Engineering, New York University. B. Huang and J. Yu are with the Department of Computer Science, Rutgers University. 
This work is supported by NSF awards IIS-1845888 and IIS-2132972; partially supported by grants ECCS-1847056 from NSF and grant W911NF-19-1-0041 from Army Research Office (ARO).
}
}

\begin{document}

\maketitle
\thispagestyle{empty}
\pagestyle{empty}

\begin{abstract}
In this study, we explore the application of game theory, in particular Stackelberg games, to address the issue of effective coordination strategy generation for heterogeneous robots with one-way communication. To that end, focusing on the task of multi-object rearrangement, we develop a theoretical and algorithmic framework that provides strategic guidance for a pair of robot arms, a leader and a follower where the leader has a model of the follower's decision-making process, through the computation of a feedback Stackelberg equilibrium. With built-in tolerance of model uncertainty, the strategic guidance generated by our planning algorithm not only improves the overall efficiency in solving the rearrangement tasks, but is also robust to common pitfalls in collaboration, e.g., chattering. 
\end{abstract}

\section{Introduction}
\label{sec:intro}

With robotic technology research and development rapidly accelerating, one can expect an explosion in the number and type of robots to be deployed in the coming years. With this trend, there is an increasing need to have robots with different capabilities effectively collaborative to solve tasks, e.g., packing products at factories or in autonomous warehouses. For example, it can be that different batches of robots have different specifications and, as a result, have complementary capabilities, which can happen when a company purchases the batches years apart. In this case, having these robots work together can effectively extend the service life of older robots, thus delivering more value for the hardware investment. However, simply putting autonomous robots together is not sufficient; algorithms must be developed to ensure that collaboration drives more value than having the robots make individual decisions. 
In the same vein, with robots increasingly permeating our work and lives, it can be predicted that robots will be working and playing alongside humans. One would expect that the robot would observe and understand human behavior and assist accordingly with limited communication.

Motivated by the above-mentioned broadly applicable use cases, in this study, we explore the application of game theory, in particular Stackelberg games \cite{von2010market}, for enabling heterogeneous autonomous robots to collaboratively solve manipulation tasks.
Specifically, we develop a framework, Stackelberg Guided Collaborative Manipulation (SGCM), for coordinating two robot arms to jointly solve a multi-object rearrangement task. The two robots have different manipulation and computation capabilities, where one is a leader and the other is a follower. The leader is assumed to have knowledge of the follower's decision-making model, whereas the follower only makes decisions based on the leader's action. SGCM is shown to deliver more efficient solutions as compared with a greedy baseline and avoids potential pitfalls, e.g., chattering where the robots nullify each other's actions, with limited communication. In other words, SGCM provides a more resilient architecture. 

This work's key contribution is a theoretical and algorithmic framework that applies Stackelberg games to robot collaboration. 
First, we propose a novel stochastic Stackelberg game framework (SGCM) to provide strategic guidance for heterogeneous robots, or agents in general, to collaboratively solve physical tasks with only leader-to-follower communication. 
Then, we developed a general algorithm, through dynamic programming and mixed integer programming, that computes the feedback Stackelberg equilibrium as the equilibrium policy for a leader-follower setting where the leader has a model of the follower's decision-making logic. 
Our algorithmic solution is instantiated and evaluated over a product-packing-like rearrangement task, which shows that the SGCM approach enables the involved robots to work together more efficiently and is robust to uncertainties. 

\begin{figure}
    \centering
    \begin{subfigure}[b]{0.2\textwidth}
        \centering
        \includegraphics[height=3.5cm]{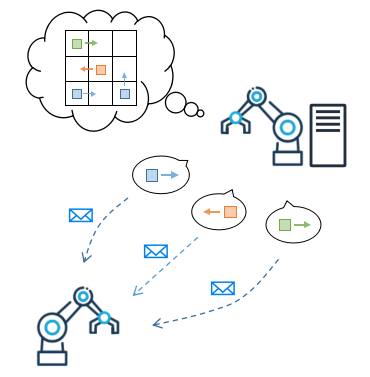}
    \end{subfigure}
    \hspace{5mm}
    \begin{subfigure}[b]{0.2\textwidth}
        \centering
        \includegraphics[height=3.5cm]{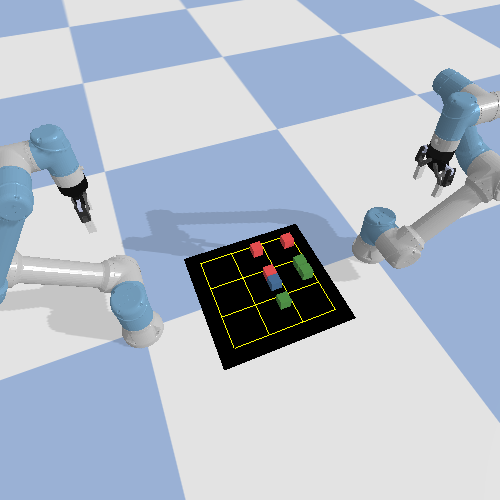}
     \end{subfigure}
    \captionsetup{belowskip=-12pt}
    \caption{Overview of Stackelberg Guided Collaborative Manipulation framework and multi-object rearrangement task settings. [left] Illustration of the setup, where one robot (upper right) has more computing power and physical capabilities would guide the other less capable robot (lower left). [right] The simulation environment in PyBullet~\cite{coumans2021}.}
    \vspace{-2mm}
    \label{fig:overview}
\end{figure}

\section{Related Work}
\label{sec:intro.related_work}

Stackelberg games are first proposed to study hierarchical competitions in the market where some companies possess dominant power \cite{von2010market}. In general, two players are involved in a Stackelberg game, a leader (she) and a follower (he). The leader first announces her strategy to maximize her utility by considering all possible reactions from the follower and sticks to that strategy. The follower best responds to the leader's strategy by maximizing his utility. Two strategies then constitute a \emph{Stackelberg equilibrium}. 
The static Stackelberg game has been extended to dynamic contexts. One direct extension is Stochastic Stackelberg Game (SSG) \cite{bacsar1998dynamic,bensoussan2015maximum}. SSG takes into consideration the feature of dynamic interactions. Therefore, the equilibrium is designed for the overall interaction process instead of a single stage. Many works have focused on SSG and its applications, for example, supply chain competitions \cite{demiguel2009stochastic}, security and resource allocation \cite{albarran2019stackelberg,tambe2011security}, and cooperative advertising \cite{bensoussan2019feedback,he2008cooperative}.

Stackelberg games are also popular in robotics research. Stankov{\'a} et al. in \cite{stankova2013staco} have proposed a Stackelberg game based approach for heterogeneous robotic swarm coverage problem, which outperforms the standard Lloyd's algorithm. Duan et al. have investigated the robot surveillance problem over graphs in \cite{duan2021stochastic}. The optimal interception strategy is studied by formulating the interaction between the intruder and the robot as a Stackelberg game. Hebbar and Langbort have discussed a collaborative human-robot search-and-rescue problem as a rescuer-rescuee Stackelberg game \cite{hebbar2020stackelberg}. The optimal rescue plans are provided with the Stackelberg game framework.
However, only a few works focus on the dynamic Stackelberg game in robotics. For example, Koh et al. in \cite{koh2020cooperative} have studied the bi-robot cooperative object transportation by formulating the problem as an SSG and applying Q-learning to solve the transportation strategy.

This work explores a new application of Stackelberg games to robotics, namely collaborative manipulation, a practical, high-utility task that finds a great many of real-world use cases. In robotics, much effort has been devoted to rendering multi-arm planning more efficient. 
In terms of general methods, 
Cohen et al. \cite{cohen2014single} adapts a heuristic search, e.g., A$^*$, in an innovative manner for solving high dimensional planning problems, including dual-arm systems. 
Shome et al. \cite{shome2020drrt} proposed the dRRT$^*$ algorithm that computes asymptotically optimal solutions that applies to multi-arm manipulation tasks. 
From an application perspective, the task of generating efficient plans for transporting objects using coordinated multi-arm rearrangement is tackled in \cite{10.1007/978-3-030-66723-8_15}. 
Xian et al. \cite{xian2017closed} proposed new techniques for performing mode-switching in solving dual-arm closed-chain manipulation tasks. 
A two-arm rearrangement task, somewhat similar to the problem examined in this work, is systematically studied in \cite{shome2021fast}. 
To our knowledge, existing works on multi-object arrangement by multiple arms generally assume central planning, which limits their application; we do not in this study.

\section{Problem Formulation}
\label{sec:prob_formulation}

We consider two robotic agents $A$ (the leader, she) and $B$ (the follower, he) cooperatively rearranging different types of objects in a 2D workspace $\Omega$. The workspace $\Omega$ is partitioned into sub-cells shown in Fig.~\ref{fig:overview}. 
Due to the heterogeneity of different robotic agents, we assume that the two robotic agents differ in their manipulation and computation capabilities. Apart from having more feasible actions to manipulate the objects, the leader also possesses more computational resources to deal with complex scenario plannings. The follower, however, has fewer feasible actions to move the objects. His computational power is also limited: he can only sense and process the current situation instead of planning for the future, or only execute the pre-programmed policies.
In addition to accomplishing the rearrangement task cooperatively, the leader can also plan ahead and guide the follower with her powerful computational resources, so that the follower achieves better utility. The full cooperation scheme ,including the guidance, can be formulated as a finite horizon Stochastic Stackelberg Game (SSG). The SGCM framework aims to provide the equilibrium policy of SSG, which is adopted for cooperation and strategic guidance in the rearrangement task. 

We define $s \in \mS$ as the environment state representing the positions of all objects in different cells of $\Omega$. $\mS$ is the set of all states. The action $a^i \in \mA^i$ represents a specific action for robot $i = \{A,B\}$ from its action set. Each action corresponds to moving one object from one cell to another. In particular, $a^i = \varnothing$ represents no action. 
We denote $T$ as the interaction horizon of SSG (also leader's prediction horizon) and use subscripts $t \geq 0$ to represent the stage. We assume both robots have a perfect observation of the current state. The game is played as follows.
\begin{itemize}[leftmargin=3mm]
    \item At stage $t < T-1$, both robots observe $s_t$. The leader first chooses her action $a^A_t \in \mA^A$. With probability $p^A_{\mathrm{fail}}$, the leader fails to execute $a^A_t$, resulting in an empty action. Then the follower reacts to $a^A_t$ by taking the action $a^B_t \in \mA^B$. With probability $p^B_{\mathrm{fail}}$, the follower fails to execute $a^B_t$, leading to an empty action. Then both robots receive the utility $u^i(s_t,a^A_t, a^B_t)$ for $i = \{A, B\}$. The environment transits to a new state $s_{t+1}$ with the the transition probability $p(s_{t+1} \vert s_t, a^A_t,a^B_t)$. 
    
    \item 
    $\cdots \cdots$
    
    \item At stage $T-1$, both robots observes $s_{T-1}$. The leader and the follower take actions $a^A_{T-1}$ and $a^B_{T-1}$ sequentially. Failure probabilities for action execution are defined similarly. The environment transits to the new state $s_T$ based on the transition probability $p(s_{T} \vert s_{T-1}, a^A_{T-1}, a^B_{T-1})$. Apart from the utility $u^i_{T-1}(s_{T-1}, a^A_{T-1}, a^B_{T-1})$, an additional terminal utility $u^i_T(s_T)$ is also incurred for robot $i = \{A,B\}$.  
\end{itemize}

\begin{remark}
The interaction horizon $T$ shows the leader's planning consideration in the cooperative rearrangement task. The leader can predict the next $T$ stages, but the rearrangement task does not necessarily terminate after $T$ stages.
When $T=1$, both robots only care about the current state, and SGCM reduces to a repeated static Stackelberg game. When $T > 1$, the leader will consider the impact from the future and compute strategies to maximize the overall utility over $T$ stages. 
\end{remark}

\begin{remark}
In the rearrangement task, since we assume $p^i_{\mathrm{fail}}$ for execution, the transition probability $p(s_{t+1} \vert s, a^A_t, a^B_t)$ is not necessarily binary given the action pair $(a^A_t, a^B_t)$. There are four possibilities for the future state $s_{t+1}$, which corresponds to the action pair $(a^A_t, a^B_t), (a^A_t, \varnothing), (\varnothing, a^B_t)$, and $(\varnothing, \varnothing)$ in the deterministic scenario. We assume $p^i_{\mathrm{fail}}$ are independent for $i=\{A,B\}$, so that the transition probabilities can be computed via $p^i_{\mathrm{fail}}$.
\end{remark}

A policy is a state-dependent probability distribution over the action set for each robot. Given state $s$, we write $\pi^A_t(a^A \vert s)$ and $\pi^B_t(a^B \vert s, \pi^A_t)$ to represent the probability\footnote{ The follower's policy is paratermized by the leader's policy in the Stackelberg equilibrium. So we include $\pi^A_t$ into $\pi^B_t$.} of choosing each action for robots $A,B$. For simplicity, we denote $\pi^i_t$ as the vector form of the robot $i$'s policy at time $t$, and write $\pi^i:= \{\pi^i_t\}_{t=0}^{T-1}$ for $i= \{A,B\}$. 
The accumulated utility for robot $i = \{A, B\}$ is given by 
\begin{equation*}
    J^i(\pi^A, \pi^B) = \E_{\pi^A, \pi^B}[\gamma^T u_T^i \vert s_0] + \sum_{t=0}^{T-1} \E_{\pi^A_{t}, \pi^B_{t}} \left[ \gamma^t u_t^i \vert s_0 \right],
\end{equation*}
where $\gamma$ is the discount factor. By taking advantage of the stage-wise additive utility and the perfect state observation, we compute the \emph{feedback Stackelberg equilibrium} (FSE) as the solution for cooperation and guidance in the rearrangement task. In the FSE $\{\pi^{A*}_t, \pi^{B*}_t \}_{t=0}^{T-1}$, at any stage of the game, both robots maximize their current and future accumulated utilities starting from the current stage.

\subsection{Guidance in Rearrangement Tasks}
\label{sec:prob_formulation.guidance}

The guidance aims to improve the follower's utility by taking advantage of the leader's powerful computation capabilities. It is embodied in two aspects: the leader plans the future for the follower and recommends the equilibrium policy to the follower. The follower can only maximize the current stage utility, but the game does not necessarily terminate in one step. Solely focusing on one-stage utility may not be optimal for the follower in the long run. However, the leader can consider the future impact and generate better strategies that are beneficial for the future. The leader can also recommend a better strategy to the follower, which cannot be computed with the follower's limited computation capability.

In SSG, one way to achieve successful guidance is to set the leader's utility the same as the follower's. In this way, as the leader maximizes her utility, she also helps maximize the follower's utility. So the equilibrium strategy is beneficial for both leader and follower. 
Another approach is to set the follower's utility as the potential function and construct the leader's utility-based the potential function, for example, affine transformation. In this way, the equilibrium strategy computed by the leader is also most beneficial for the follower.

\section{Stackelberg Guided Collaborative Manipulation Framework}
\label{sec:dp_milp}

In this section, we use dynamic programming to solve the FSE policies for both robots. In the equilibrium computation, we reformulate the leader's problem into a Mixed Integer Linear Programming (MILP) to reduce the problem complexity. Next, we propose the Stackelberg Guided  Collaborative Manipulation (SGCM) framework to compute cooperation and guidance strategies for both robots with a rolling horizon approach and summarize the algorithm.

\subsection{Dynamic Programming for Computing FSE}
\label{sec:dp_milp.dp}

The FSE generates an optimal cooperation strategy that utilizes the leader's computational advantage. The leader can plan for $T$ stages to envision the optimal action to reorganize the objects and the optimal guidance strategy to benefit the follower. The FSE can be solved by dynamic programming retrospectively. 

We define $v_t^i(s)$ as the value function for the robot $i$ at stage $t$, $i \in \{A,B\}$. The terminal value $v_T^i(s)$ is given by the terminal utility $u^i_T(s_T)$. At each stage of the game, two robots play a Stackelberg game with perfect state observation. At stage $t \leq T-1$, after observing the state $s_{t}$, the $t$-th component of the equilibrium is computed by 
\begin{equation}
\label{eq:QA_t}
\tag{$\mQ^A_t$}
\begin{split}
    \max_{\pi_{t}^A} \quad & \sum_{a^A, a^B} \pi_{t}^{B^*}(a^B \vert s, \pi^A_{t}) \pi^A_{t}(a^A \vert s) \\
    & \qquad \left[ u^A_t(s, a^A, a^B) + \sum_{s'} p(s' \vert s, a^A, a^B) v_{t+1}^A(s') \right] \\ 
    \text{s.t.} \quad & 0 \leq \pi^A_{t}(a^A \vert s) \leq 1, \quad \forall a^A \in \mA^A, \\ 
    & \sum_{a^A} \pi^A_{t}(a^A \vert s) = 1, \\ 
    & \pi^{B*}_{t} \in \arg\max_{\pi^B_{t}} \ \mQ^B_t(\pi^A_{t}).
\end{split}
\end{equation}
where  
\begin{equation}
\label{eq:QB_t}
\tag{$\mQ^B_t(\pi^A_t)$}
\begin{split}
    \max_{\pi_{t}^B} \quad & \sum_{a^A, a^B} \pi^B_{t}(a^B \vert s, \pi^A) \pi_{t}^A(a^A \vert s) \\
    & \qquad \left[ u^B_t(s, a^A, a^B) + \sum_{s'} p(s' \vert s, a^A, a^B) v_{t+1}^B(s')\right] \\ 
    \text{s.t.} \quad & 0 \leq \pi^B_{t}(a^B \vert s, \pi^A) \leq 1, \quad \forall a^B \in \mA^B, \\ 
    & \sum_{a^B} \pi^B_{t}(a^B \vert s, \pi^A) = 1.
\end{split}
\end{equation}
For simplicity, we write $\sum_{a^A \in \mA^A}$ as $\sum_{a^A}$. The same applies to $\sum_{a^B}$. The summation over $s'$ contains four possibilities as discussed in Sec.~\ref{sec:prob_formulation}.
We assume the optimal solution set of \eqref{eq:QB_t} is a singleton for all $t = 0,\dots, T-1$. Then $\pi^{B*}_t(a^B \vert s, \pi^A_t)$ is unique given the leader's policy $\pi^A_t$. 
After solving for $\pi^{A*}_t$ and $\pi^{B*}_t$, we update the value function $v^i_t(s)$, $i = \{A,B\}$, by setting them as the optimal objective values of \eqref{eq:QA_t} and \eqref{eq:QB_t}. The FSE can be obtained by solving \eqref{eq:QA_t} and \eqref{eq:QB_t} backward from stage $T-1$ to stage $0$. 

\subsection{MILP Reformulation}
\label{sec:dp_milp.milp}

To find the $t$-th component of the FSE, we need to solve a bilevel optimization problem \eqref{eq:QA_t}, which is in general hard. However, we notice that the problem \eqref{eq:QB_t} is an LP and is linear in $\pi^B_t$. By utilizing this structure, we can use KKT conditions to equivalently represent \eqref{eq:QB_t}. Note that the constraint $\pi_{t}^B(a^B \vert s, \pi^A) \leq 1$ is in fact redundant because it is guaranteed by $\pi_{t}^B(a^B \vert s, \pi^A) \geq 0$ and $\sum_{a^B} \pi_{t}^B(a^B \vert s, \pi^A) = 1$. Therefore we omit this inequality and simplify \eqref{eq:QB_t} as 
\begin{equation}
\label{eq:tilde_QB_t}
\tag{$\tilde{\mQ}^B_t(\pi^A_t)$}
\begin{split}
    \max_x \quad & (\pi^A_t)^\tp U^B_t \pi^B_t \\
    \text{s.t.} \quad & \pi^B_t \geq 0, \quad \sum_{a^B} \pi^B_t = 1,
\end{split}
\end{equation}
where the utility matrix $U^B_t \in \R^{\abs{\mA^A} \times \abs{\mA^B}}$ and $U^B_{t,ij} = u^B_t(s, a^A, a^B) + \sum_{s'} p(s' \vert s, a^A, a^B) v_{t+1}^B(s')$. 

Let $\lambda \in \R$ and $\mu \in \R^{\abs{\mA^B}}$ be the dual variables associated with the equality and the inequality constraints in \eqref{eq:tilde_QB_t}, respectively. We obtain the KKT conditions
\begin{equation}
\label{eq:follower_kkt}
\begin{aligned}
    &\lambda \1_{\abs{\mA^B}} - (U_t^B)^\tp \pi^A_t \geq 0, && \pi^B_t \geq 0, \\ 
    &(\pi^B_t)^\tp (\lambda \1_{\abs{\mA^B}} - (U_t^B)^\tp \pi^A_t) = 0, && \sum_{a^B} \pi^B_t = 1,
\end{aligned}
\end{equation}
where $\1_{\{\cdot\}}$ is the all-ones vector with proper dimensions.

For a Stackelberg game, the pure strategy for the follower always exists, which means that $\pi^B_t$ can only have one non-zero element. If $\pi^{B*}_t$ of \eqref{eq:tilde_QB_t} is not a singleton, we can show that any pure strategy in the support of $\pi^{B*}_t$ is also optimal \cite{paruchuri2008playing}. Therefore, we only focus on the pure strategy for the follower, which can be represented by binary variables. Furthermore, we can use the binary variable to linearize the complementarity condition in \eqref{eq:follower_kkt} and obtain 
\begin{equation*}
    0 \leq \lambda \1 - (U_t^B)^\tp \pi^A_t \leq M (1-\pi^B_t), \quad \pi^B_t = \{0, 1\}^{\abs{\mA^B}},
\end{equation*}
where $M$ is a large number. We substitute the inner optimization problem in \eqref{eq:QA_t} with KKT conditions and obtain 

\begin{equation}
\label{eq:tilde_QA_t}
\tag{$\tilde{\mQ}^A_{t}$}
\begin{split}
    \max_{\pi_{t}^A, \pi_{t}^B, \lambda_{t}} \quad & (\pi^A_t)^\tp U^A_t \pi^B_t \\ 
    \text{s.t.} \quad & 0 \leq \pi^A_{t} \leq 1, \quad \qquad \sum_{a^A} \pi^A_{t} = 1, \\
    & \pi^B_{t} \in \{0, 1\}^{\abs{\mA^B}}, \quad \sum_{a^B} \pi^B_{t} = 1, \\ 
    & 0 \leq \lambda_{t} \1_{\abs{\mA^B}} - (U_t^B)^\tp \pi^A_{t} \leq M( 1-\pi^B_{t}),
\end{split} 
\end{equation}
where the utility matrix $U^A_t \in \R^{\abs{\mA^A} \times \abs{\mA^B}}$ and $U^A_{t,ij} = u^A_t(s, a^A, a^B) + \sum_{s'} p(s' \vert s, a^A, a^B) v_{t+1}^A(s')$. 

We note that \eqref{eq:tilde_QA_t} is a mixed integer quadratic programming (MIQP). To facilitate the computation, we follow \cite{paruchuri2008playing} to further cast \eqref{eq:tilde_QA_t} to an MILP by changing of variables $z_{t}(a^A, a^B) = \pi_{t}^A(a^A \vert s) \pi_{t}^B(a^B \vert s, \pi^A_{t})$. Then $z_t \in \R^{\abs{\mA^A} \times \abs{\mA^B}}$ and \eqref{eq:tilde_QA_t} becomes
\begin{equation}
\label{eq:bar_QA_t}
\tag{$\bar{\mQ}^A_t$}
\begin{split}
    &\max_{z_{t}, \pi_{t}^B, \lambda_{t}} \quad  U^A_t \odot z_t \\ 
    &\begin{aligned}
        \text{s.t.} \ & \pi^B_{t} \in \{0, 1\}^{\abs{\mA^B}}, \quad \sum_{a^B} \pi^B_{t} = 1, \\ 
        & 0 \leq z_t \leq 1, \quad \sum_{a^A, a^B} z_{t} = 1, \quad z_{t} \1_{\abs{\mA^B}} \leq \1_{\abs{\mA^A}}, \\
        & \pi_{t}^B \leq z_{t}^\tp \1_{\abs{\mA^A}} \leq \1_{\abs{\mA^B}}, \\
        & 0 \leq \lambda_{t} \1_{\abs{\mA^B}} - (U^B_t)^\tp \left( z_{t} \1_{\abs{\mA^B}} \right) \leq M( 1-\pi^B_{t} ),
    \end{aligned}
\end{split} 
\end{equation}
where $\odot$ represents element-wise multiplication. Since $z_t$ is a matrix variable, we use $\sum_{a^A, a^B} z_t$ to denote the sum of all the element in $z_t$.

\begin{remark}
The finite horizon game allows us to define $T+1$ subsets of $\mS$ given the initial state $s_0$: $\{ \mS_0, \mS_1, \dots, \mS_T \} := \{\mS_t\}_{t=0}^T$ and $\mS_0 = \{s_0\}$. Each $\mS_t$ contains all possible states occurred at time $t$. The value function $v_t^i(s)$ $(i=\{A, B\})$ is only defined for the state $s \in \mS^t$ instead of all the states in $\mS$. In this way, we do not need to compute $v^i_t(s)$ for state $s \not\in \mS_t$, $i=\{A, B\}$, which saves computation time.
It is clear that $\mS_t \subset \mS$ for $t=0,\dots, T$, but we do not necessarily have $\mS_m \cap \mS_n = \varnothing$ for $m \neq n$. This means that the interactions between two robots may lead to some old states.
\end{remark}

\subsection{Rolling Horizon Computation for Rearrangement Task}
\label{sec:dp_milp.mhc}

We mention that the cooperative rearrangement task does not necessarily terminate after $T$ stages. Therefore, we propose the SGCM framework which adopts the rolling horizon approach to compute FSE and execute the first-stage FSE policy. Similar to Model Predictive Control (MPC), the rolling horizon approach improves the robustness and resiliency of the cooperation. 
The follower may not precisely execute the leader's recommended strategy if he is susceptible to uncertainties. 
The leader can readjust her action and guidance policy accordingly to minimize the impact of uncertainties. The task eventually terminates when all objects are reorganized to the goal position, which corresponds to $s_{\mathrm{goal}}$.
We summarize the SGCM framework for cooperative rearrangement with strategic guidance in Alg.~\ref{alg:1}.

\vspace{-1mm}

\begin{algorithm}[h]
\begin{small}
\KwInit $s_0, s_{\mathrm{goal}}, T$ \;
\For{iter = $1,2,\dots$}{
    \tcp{Forward prediction}
    $t \gets 0$, $\mS_0 = \{s_0\}$\;
    \While{$t < T$}{
        \For{$s \in \mS_t$}{
            \For{action pair $(a^A, a^B) \in \mA^A\times \mA^B$}{
                predict $s_{\mathrm{new}}$ with $s, (a^A, a^B)$ \;
                add $s_{\mathrm{new}}$ to $\mS_{t+1}$ \;
                compute and store $u^i(s, a^A, a^B)$, $i = \{A, B\}$ \;
            }
        }
        $t \gets t+1$ \;
    }
    \tcp{Dynamic programming}
    $v^i_T(s) \gets u^i_T(s) \ \forall s \in \mS_T$, $i = \{A, B\}$ \;
    $t \gets T-1$ \;
    \While{$t \geq 0$}{
        \For{$s \in \mS_t$}{
            formulate utility matrices $U^A_t, U^B_t$ \;
            $(\pi^{A*},\pi^{B*}) \gets$ solve MLIP \eqref{eq:bar_QA_t} \;
            $v^i_t(s) \gets (\pi^{A*})^\tp U_t^i \pi^{B*}$, $i = \{A,B\}$ \;
        }
        $t \gets t - 1$ \;
    }
    \tcp{Game execution and observation}
    leader executes $\pi^{A*}_0$ and recommend $\pi^{B*}_0$ \;
    observe new state $s_{\mathrm{obs}}$\;
    \If{$s_{\mathrm{obs}} = s_{\mathrm{goal}}$}{
        \kWBreak \;
    }
    $s_0 \gets s_{\mathrm{obs}}$ \;
}
\caption{SGCM framework}
\label{alg:1}
\end{small}
\end{algorithm}

\vspace{-1mm}
\begin{remark}
In Alg.~\ref{alg:1}, for problems with small size, we can always enumerate all possible states in the forward prediction and perform value iterations to compute the FSE policy as shown in the algorithm. We mention that other simulation-based approaches can also be easily incorporated in Alg.~\ref{alg:1} if the problem size is large. For example, we can adopt Monte Carlo Tree Search (MCTS) to explore part of $\mS_t$ at different stage $t$ if the size of $\mS_t$ is huge. Then we only perform value iterations on the simulated states and compute approximate FSE policy. The simulation-based approaches are not guaranteed to provide the global optimal equilibrium, although they may be faster for online computation. However, for a specific task such as the rearrangement task, we can always pre-process states. Then we only need to perform a state search in the forward prediction for online computation. 
\end{remark}

\section{Experiments and Evaluations}
\label{sec:exp_eval}

In this section, we evaluate our SGCM framework and demonstrate the strategic guidance with a multi-object rearrangement task \cite{han2018complexity}, where two heterogeneous robotic arms (also called robots) cooperatively rearrange the objects to the goal position. The basic settings are shown in Fig.~\ref{fig:overview}. Two robots are distinguished by their manipulation and computation capabilities. The leader can move the object along horizontal, vertical, and diagonal directions; she is also capable of making complex planning and predictions to accomplish the task by sensing the environment. The follower, however, can only manipulate the objects along the horizontal and vertical directions. His limited computation capability only allows him to consider the current situation rather than the future.
We represent objects of different types with different colors for visualization purposes. The goal is to rearrange the red, green, and blue objects to bottom-left, bottom-middle, and bottom-right cells, respectively. Every action of the two robots is assigned a cost, and the cost of manipulating the object from a specific cell may double, depending on how many objects are in that cell. 
The state is defined in Sec.\ref{sec:prob_formulation}. At the beginning of each round of interaction, a reward is assigned to the current state, which is proportional to the distance of the current state to the goal state. Each robot's utility is the reward minus the cost.

The experiment is carried out by simulation (PyBullet~\cite{coumans2021}), and the perception to the environment is vision-based. We assume that the ground truth segmentation of objects is accessible in the simulation, which is reasonable due to the practicality in real-world object detection methods such as Mask R-CNN~\cite{he2017mask}. 
The overall system works as follows. The leader perceives the environment state (positions of all objects) to compute high-level object manipulation command: what to grasp and where to place. After receiving the command, a low-level controller executes the command. The low-level controller adopts a Grasp Network~\cite{HuaHanBouYu21ICRA,huang2021visual} to propose the grasp position of the selected object. Then the PyBullet’s internal inverse kinematics module is used for manipulation motion planning. The position for placement is obtained by pixel test, where all pixels in the destination cells are iteratively tested until a pixel is found such that the object can be placed at this pixel as the center and has no collisions with other objects.

\subsection{SGCM Framework and Greedy Approach}
\label{sec:exp_eval.1}

\begin{figure}[h!]
\vspace{-3mm}
    \centering
    \begin{subfigure}[b]{0.5\textwidth}
        \centering
        \includegraphics[height=1.68cm]{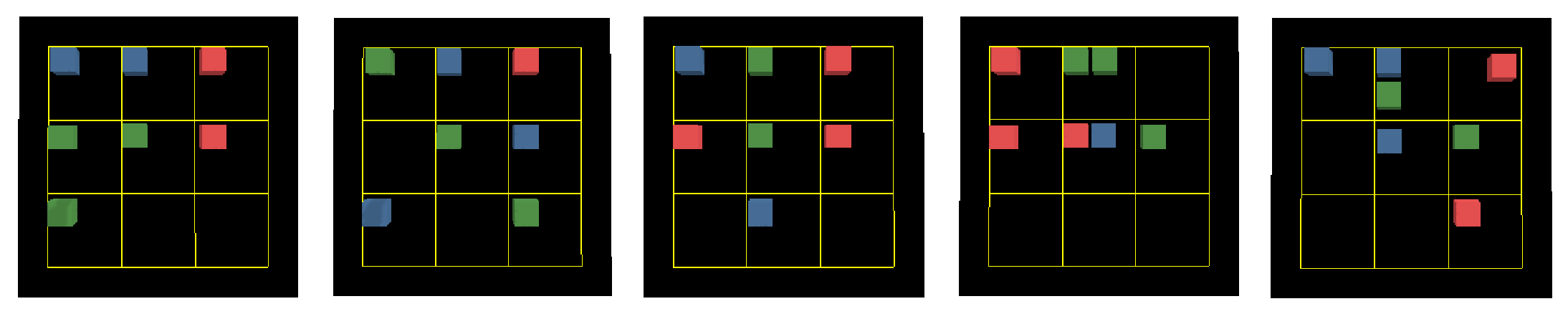}
        \captionsetup{aboveskip=0pt}
    \end{subfigure}

    \begin{subfigure}[b]{0.5\textwidth}
        \centering
        \includegraphics[height=1.68cm]{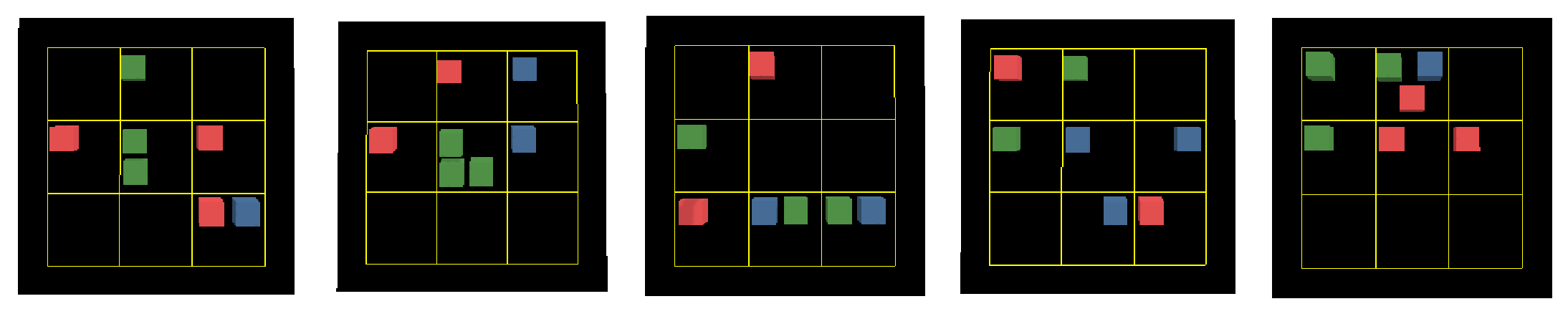}
        \captionsetup{aboveskip=0pt}
     \end{subfigure}
\vspace{-5mm}
    \caption{The initial configurations of 10 cases in the experiment.}
\vspace{-3mm}
    \label{fig:initial_config}
\end{figure}

We set the interaction horizon $T=2$ and assume the failure probabilities $p^A_{\mathrm{fail}} = p^B_{\mathrm{fail}}=0.1$. Then two robots follow the SGCM framework to cooperatively work on the rearrangement task. For comparison, we also implemented a greedy rearrangement strategy. In the greedy approach, there is in fact no cooperation between two robots. Each robot observes the current environment and moves sequentially, only maximizing his/her current-stage utility. 
The reason for using the one-stage utility is because of no cooperation. Each robot has no need to consider the impact of the other robot. 
We select $10$ different cases for testing. Each case corresponds to a different initial object setting. In these cases, the number of objects in the same type may differ, but the goal positions do not change. The initial configurations of 10 cases are shown in Fig.~\ref{fig:initial_config}.

We summarize the experiment results of using the SGCM framework and greedy approaches in Tab.~\ref{tab:comparison}. 
\vspace{-2mm}

\begin{table}[htbp]
    \centering
    \begin{tabular}{c|ccc|ccc}
    \hline
        & \multicolumn{3}{c|}{Greedy} & \multicolumn{3}{c}{SGCM} \\ \hline
        case \# & status & rounds & utility & status & rounds & utility \\ \hline 
        1 & No  & $>7$  & 283.5 & Yes & 6 & 309 \\ \hline
        2 & No  & $>7$  & 288.5 & Yes & 6 & 313.5 \\ \hline
        3 & No  & $>7$  & 300.5 & Yes & 6 & 315 \\ \hline
        4 & Yes & 5     & 278   & Yes & 5 & 283 \\ \hline
        5 & No  & $>8$  & 306.5 & Yes & 7 & 344.5 \\ \hline
        6 & Yes & 6     & 289.5 & Yes & 5 & 306.5 \\ \hline
        7 & Yes & 5     & 297.5 & Yes & 5 & 298.5 \\ \hline
        8 & Yes & 3     & 219.5 & Yes & 3 & 219.5 \\ \hline
        9 & Yes & 7     & 249   & Yes & 5 & 275 \\ \hline
        10 & Yes & 7    & 277.5 & Yes & 6 & 301 \\ \hline
    \end{tabular}
    \caption{Comparison between SGCM framework and greedy approach for 10 cases. Status refers to the completion of the task. The utility is the sum of single-stage utility over interaction rounds. }
    \vspace{-3mm}
    \label{tab:comparison}
\end{table}

Note that robots do not care about cooperation in the greedy approach. 
To measure the performance of the greedy approach and to compare with the SGCM framework, we define the stage-wise utility in the greedy approach as the current state reward minus the total cost of two robots after each round of manipulation. The utility in Tab.~\ref{tab:comparison} is the sum of all stage-wise utility and the values are comparable. 
Also, note that the SGCM framework and the greedy approach may have different interaction rounds. For comparison, the utility in Tab.~\ref{tab:comparison} is computed based on the rounds from the SGCM framework because the SGCM framework always yields a smaller one. 
It means that we only sum up the single-stage utility in the greedy approach before the specified rounds, regardless of the completion status.

\begin{figure}[h]
    \vspace{-3mm}
    \begin{subfigure}[b]{0.45\textwidth}
        \centering
        \includegraphics[height=1.55cm]{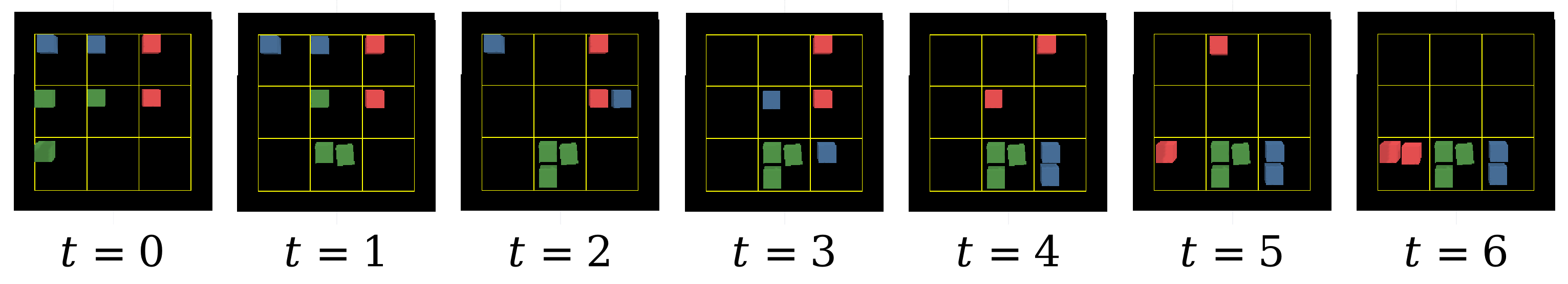}
        \captionsetup{aboveskip=0pt}
        \caption{Rearrangement status at each stage with SGCM framework.}
        \label{fig:task_evolution.1}
    \end{subfigure}
    
    \begin{subfigure}[b]{0.45\textwidth}
        \centering
        \includegraphics[height=1.55cm]{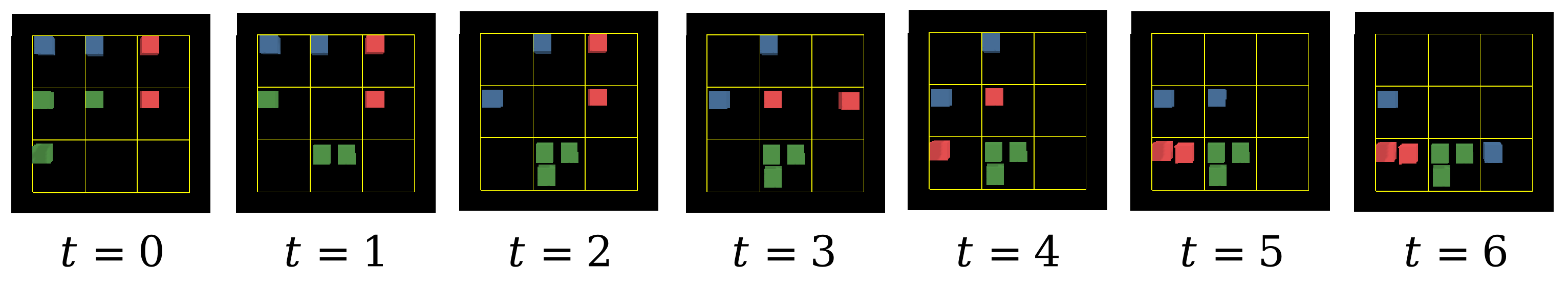}
        \captionsetup{aboveskip=0pt}
        \caption{Rearrangement status at each stage with greedy approach.}
        \label{fig:task_evolution.2}
    \end{subfigure}
        \captionsetup{belowskip=-10pt}
    \caption{Interaction evolution for Case 1 using SGCM framework and greedy approach. SGCM framework completes the task after 6 steps, while the greedy approach fails to rearranges objects and gets stuck in the last state.}
    \label{fig:task_evolution}
\end{figure}

An example of the evolution of the interactions in the rearrangement task is visualized in Fig.~\ref{fig:task_evolution}.

From Tab.~\ref{tab:comparison}, we can observe several advantages of our SGCM framework over the greedy approach in the cooperative rearrangement task:
\begin{itemize}[leftmargin=3mm]
    \item 
    Two robots using the greedy approach can get stuck in some states due to myopic strategies. In these states, two robots repeat single actions, and the objects will never be reorganized to the target position.
    The SGCM framework can avoid such situations by taking advantage of the leader's computation capabilities. It also shows the significance of the planning and strategic guidance to the integrity of the cooperative rearrangement task. See Case 1, 2, 3, 5.
    
    \item
    When two robots are able to finish the rearrangement task with the greedy approach, the SGCM framework can either reduces the number of interactions and saves more actions (Case 6,9,10), or achieves higher utility when the number of interactions are the same (Case 4,7), showing the outperformance of the SGCM framework.
\end{itemize}

\begin{remark}
For some simple cases where the objects are easy to rearrange, for example, Case 8, the SGCM framework has the same performance as the greedy approach. However, this does not harm the effectiveness of the SGCM framework. In practice, we do not always have simple cases to rearrange. Then the advantage of the SGCM framework starts to appear, as demonstrated in other cases.
\end{remark}

\begin{figure}[h!]
\vspace{-1mm}
    \centering
    \begin{subfigure}[b]{0.235\textwidth}
        \centering
        \includegraphics[height=3cm]{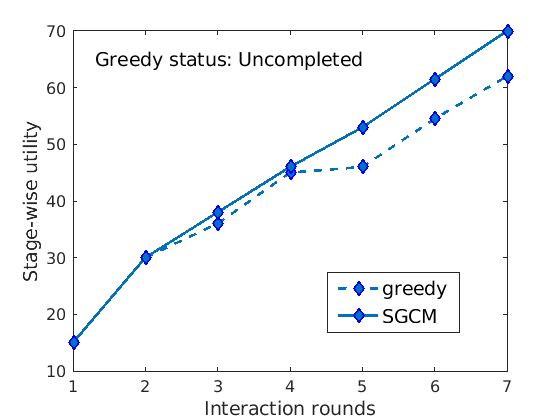}
        \caption{For test case 2.}
        \label{fig:utility.1}
    \end{subfigure}
    \hfill
    \begin{subfigure}[b]{0.235\textwidth}
        \centering
        \includegraphics[height=3cm]{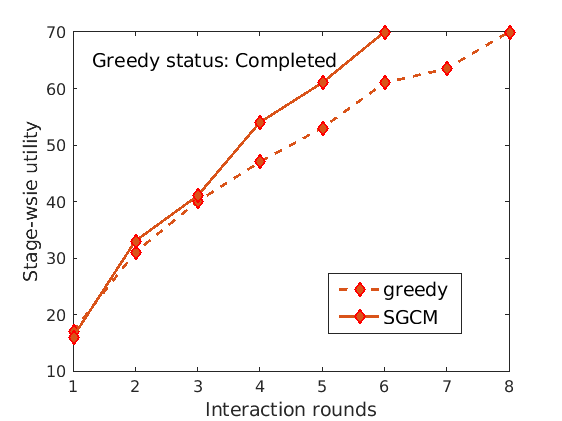} 
        \caption{For test case 9.}
        \label{fig:utility.2}
     \end{subfigure}
    \caption{Stage-wise utility for two different cases, which demonstrates that the SGCM framework outperforms the greedy approach.}
\vspace{-2mm}
    \label{fig:utility}
\end{figure}

In order to have a clear view of how the SGCM framework outperforms the greedy approach, we plot the stage-wise utility along with the interactions in Fig.~\ref{fig:utility}. We observe that at the beginning of the interaction, two approaches have the same stage-wise utility. But as the interaction evolves, the SGCM framework starts to yield a higher stage-wise utility than the greedy approach, showing the power of the strategic guidance in the rearrangement task.

\subsection{SGCM with Disturbance and Zero Trust}
\label{sec:exp_eval.2}

We demonstrate that our SGCM framework is robust and resilient to uncertainties and disturbances. In practice, the follower may not precisely execute the leader's recommended strategy at every stage due to the following reasons. First, the external disturbance may lead to hardware failure; second, the follower may not trust the leader's recommendation; third, the robot may be infected or hijacked by malware due to cybersecurity issues. In this situation, the follower may seek the strategy by himself, or randomly select an action, or even becomes adversarial to the leader. Therefore, resiliency is indispensable for cooperation. We illustrate the resiliency of our SGCM framework by injecting disturbances during the interaction, i.e., the follower does not follow the recommended strategy at certain steps. Our framework allows the leader to sense the abnormality and to adjust her strategy as well as the new recommended strategy in time, so that the impact of the disturbance is minimized. 

\begin{figure}[h!]
\vspace{-1mm}    \centering
    \begin{subfigure}[b]{0.235\textwidth}
        \centering
        \includegraphics[height=3cm]{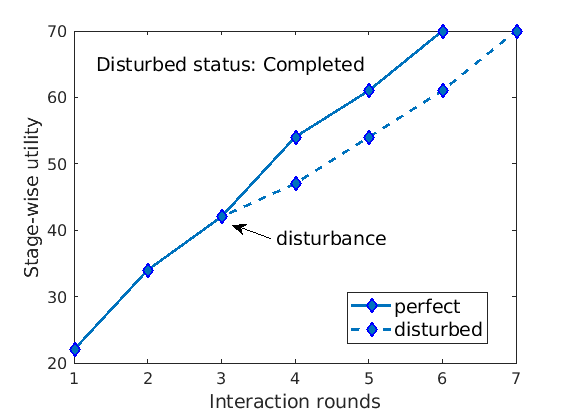}
        \caption{For test case 4.}
        \label{fig:disturb.1}
    \end{subfigure}
    \hfill
    \begin{subfigure}[b]{0.235\textwidth}
        \centering
        \includegraphics[height=3cm]{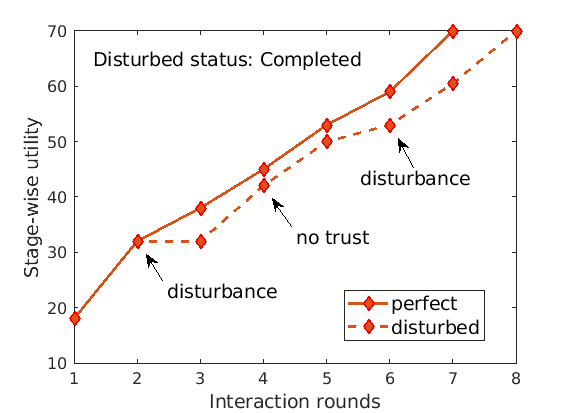} 
        \caption{For test case 3.}
        \label{fig:disturb.2}
    \end{subfigure}
    \caption{Two plots shows how the disturbance and zero trust affects the stage-wise utility for two different cases. In both cases two robots are able to complete the rearrangement, which demonstrates the resiliency and robustness of the Stackelberg game approach.}
    \label{fig:disturb}
\vspace{-2mm}
\end{figure}

In Fig.~\ref{fig:disturb}, the follower randomly selects a feasible action instead of the leader's recommended strategy to manipulate the objects when a ``disturbance" occurs. The ``no trust" in Fig.~\ref{fig:disturb.2} means that the follower does not trust the leader's recommendation and selects the greedy strategy for manipulation. It is not surprising to see the performance degeneration after the disturbance or the zero trust. However, the deviation between the perfect and disturbed cases is controlled by SGCM and does not explode. Although suffering the disturbance and trust issues, we see that the SGCM framework can still ensure two robots accomplish the rearrangement task successfully. It shows that our SGCM framework is resilient and robust to random failure and zero trust during the two-robot cooperation in the rearrangement task.

\section{Conclusion}
\label{sec:conclusion}
In this paper, we have proposed a Stackelberg Guided Collaborative Manipulation (SGCM) framework for heterogeneous robots collaboration. Focusing on the multi-object rearrangement task, the SGCM framework enables the leader robot to strategically guide the follower robot with her more powerful manipulation and companion capabilities to achieve better performance. The feedback Stackelberg equilibrium is adopted as the guidance strategy in the SGCM framework, which can be computed effectively by the developed algorithm. The SGCM framework only requires one-way communication to cooperatively work on the rearrangement task. In addition, our SGCM framework is also robust and resilient to uncertainties, disturbances, and trust issues during cooperation. Besides the theoretical guarantees, the effectiveness of the SGCM framework is thoroughly evaluated and validated over many different test cases of rearrangement tasks, where our approach displayed a number of advantages over greedy approaches. For future, we intend to extend our framework to a multi-follower setting, where the benefit of the Stackelberg approach is expected to become more prominent. More learning aspects such as learning the follower's behavior pattern will also be considered.

\addtolength{\textheight}{-11.2cm}   





\bibliographystyle{IEEEtran}
\bibliography{main.bib}

\end{document}